\documentclass[10pt, letterpaper, twocolumn, journal, final]{IEEEtran}
\usepackage{amsmath}
\usepackage[hidelinks]{hyperref}
\usepackage{amssymb}
\usepackage{cite}
\usepackage{booktabs}
\usepackage{graphicx}
\usepackage{setspace}
\usepackage{algpseudocode}
\usepackage{algorithm}
\usepackage{tikz}
\usetikzlibrary{shapes,arrows, positioning, fit}
\usepackage{epstopdf}
\usepackage{verbatim}
\let\Algorithm\algorithm
\renewcommand\algorithm[1][]{\Algorithm[#1]\setstretch{1.5}}

\algnewcommand{\inputs}[1]{%
  \State \textbf{inputs:}
  \Statex \hspace*{\algorithmicindent}\parbox[t]{.8\linewidth}{\raggedright #1}
}
\algnewcommand{\initialize}[1]{%
  \State \textbf{initialize:}
  \Statex \hspace*{\algorithmicindent}\parbox[t]{.8\linewidth}{\raggedright #1}
}
\allowdisplaybreaks
\tikzstyle{block} = [rectangle, draw, fill=lightgray!20, 
    text width=4em, text centered, rounded corners, minimum height=2em]
    \tikzstyle{line} = [draw, -latex']
    \tikzstyle{cloud} = [draw, ellipse,fill=black!20, node distance=2.5cm,minimum height=2em]

\renewcommand{\vec}[1]{\mathnormal{#1}}
\renewcommand{\leq}{\leqslant}
\renewcommand{\geq}{\geqslant}

\title{Convex Denoising using Non-Convex Tight Frame Regularization}
\author{
{Ankit Parekh and Ivan W. Selesnick}
\thanks{Copyright (c) 2015 IEEE. Personal use of this material is permitted. However, permission to use this material for any other purposes must be obtained from the IEEE by sending a request to pubs-permissions@ieee.org.}\thanks{A. Parekh (ankit.parekh@nyu.edu) is with the Department of Mathematics and I. Selesnick (selesi@nyu.edu), is with the Department of Electrical and Computer Engineering at School of Engineering, New York University, Brooklyn, NY.}\thanks{MATLAB software is available at \color{blue}\url{https://goo.gl/Wkd5wc}}}

\begin{document}
\maketitle	
\begin{abstract}
This letter considers the problem of signal denoising using a sparse tight-frame analysis prior. The $\ell_1$ norm has been extensively used as a regularizer to promote sparsity; however, it tends to under-estimate non-zero values of the underlying signal. To more accurately estimate non-zero values, we propose the use of a non-convex regularizer, chosen so as to ensure convexity of the objective function. The convexity of the objective function is ensured by constraining the parameter of the non-convex penalty. We use ADMM to obtain a solution and show how to guarantee that ADMM converges to the global optimum of the objective function. We illustrate the proposed method for 1D and 2D signal denoising.
\end{abstract}


\section{Introduction}
\label{sec::Introduction}

\PARstart{A}{} standard technique for estimating sparse signals is through the formulation of an inverse problem with the $\ell_1$ norm as convex proxy for sparsity. In particular, consider the problem of estimating a signal $\vec{x} \in \mathbb{R}^n$ from a noisy observation $\vec{y} \in \mathbb{R}^n$,
\begin{align}
\vec{y} = \vec{x} + \vec{w}, 
\end{align}where $\vec{w}$ represents AWGN. We assume the underlying signal to be sparse with respect to an overcomplete tight frame $\vec{A} \in \mathbb{R}^{m \times n}$, $m \geq n$, which satisfies the tight frame condition, i.e., 
\begin{align}
\label{eq::Parseval frame}
\vec{A}^{T}\vec{A} = r\vec{I}, \quad r > 0.
\end{align}Using an analysis-prior, we formulate the signal denoising problem as
\begin{align}
\label{eq::Cost function}
\arg\min_{\vec{x}} \Biggl \lbrace F(\vec{x}) := \dfrac{1}{2}\|\vec{y}-\vec{x}\|_2^2 + \sum_{i = 1}^{m}\lambda_i\phi\left([\vec{Ax}]_i; a_i \right)\Biggr\rbrace,
\end{align} where $\lambda_i > 0$ are the regularization parameters, and $\phi\colon\mathbb{R}\to\mathbb{R}$ is a non-smooth sparsity inducing penalty function. The parameters $a_i$ control the non-convexity of $\phi$ in case it is non-convex. The analysis prior is used in image processing and computer vision applications \cite{Selesnick2009,Elad2007, Cai2014,Cai2010,Xie2012, Turek2014,Portilla2009}. 
Commonly, the $\ell_1$ norm is used to induce sparsity, i.e., $\phi(x) = |x|$ \cite{Tropp_2006_tinfo,Chen1998}. In that case, problem \eqref{eq::Cost function} is strictly convex and the global optimum can be reliably obtained.

The $\ell_1$ norm is not the tightest envelope of sparsity \cite{Jojic2011}. It under-estimates the non-zero values of the underlying signal \cite{Nikolova1998, Candes2008}. Non-zero values can be more accurately estimated using suitable non-convex regularizers. Non-convex regularization in an analysis model has been  used for MRI reconstruction \cite{Chartrand2009}, EEG signal reconstruction \cite{Majumdar2014}, and for computer vision problems \cite{Ochs2015}. However, the use of non-convex regularizers comes at a price: the objective function is generally non-convex. Consequently, several issues arise (spurious local minima, a perturbation of the input data can change the solution unpredictably, convergence is guaranteed to the local minima only, etc.). 

In order to maintain convexity of the objective function while using non-convex regularizers, we propose to restrict the parameter $a_i$ of the non-convex regularizer $\phi$. By controlling the degree of non-convexity of the regularizer we guarantee that the total objective function $F$ is convex. This idea which dates to Blake and Zisserman \cite{Blake1987} and Nikolova \cite{Nikolova1998}, has been applied  to image restoration and reconstruction \cite{Nikolova1999,Nikolova2010}, total variation denoising \cite{Selesnick2015, Lanza2015}, and wavelet denoising \cite{Selesnick2015_2}.

In this letter we provide a critical value of parameter $a$ to ensure $F$ in \eqref{eq::Cost function} is strictly convex (even though $\phi$ is non-convex). In contrast to the above works, we consider transform domain regularization and prove that ADMM \cite{Boyd2010a} applied to the problem \eqref{eq::Cost function} converges to the global optimum. The convergence of ADMM is guaranteed, provided the augmented Lagrangian parameter $\mu$, satisfies $\mu > 1/r$.

\section{Sparse signal estimation}
\label{sec::Sparse Signal estimation}

\subsection{Non-convex Penalty Functions}
\label{subsec::Penalty Functions}
In order to induce sparsity more strongly than the $\ell_1$ norm, we use non-convex penalty functions $\phi\colon\mathbb{R}\to\mathbb{R}$ parameterized by the parameter $a \geq 0$. We make the following assumption of such penalty functions.

\newtheorem{assumption}{\bf Assumption}
\begin{assumption}
\label{theorem::assumption 1}
The non-convex penalty function $\phi\colon\mathbb{R} \to \mathbb{R}$ satisfies the following
\begin{enumerate}
\item $\phi$ is continuous on $\mathbb{R}$, twice differentiable on $\mathbb{R}\!\setminus\! \lbrace 0\rbrace$ and symmetric, i.e., $\phi(-x; a) = \phi(x; a)$
\item $\phi'(x) > 0, \forall x > 0$	
\item $\phi''(x) \leq 0, \forall x > 0$
\item $\phi'(0^{+}) = 1$
\item $\inf\limits_{x\neq0}\phi''(x;a) = \phi''(0^+;a) = -a$
\item $\phi(x;0) = |x|$.
\end{enumerate}
\end{assumption}

\begin{figure}
\centering
{}\includegraphics[]{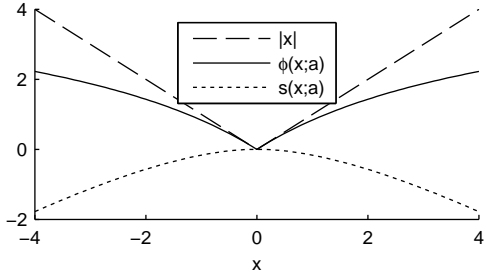}
\caption{ The non-differentiable rational penalty function $\phi(x;a)$ and the function $s(x;a) = \phi(x;a) - |x|$, $a = 0.4$.}
\label{fig::phi minus absolute}
\end{figure}

Since $\phi(x;0) = |x|$, the $\ell_1$ norm is recovered as a special case of the penalty function $\phi$. The parameter $a$ controls the degree of non-convexity of $\phi$. Note that the $\ell_p$ norm does not satisfy assumption \ref{theorem::assumption 1}. The rational penalty function \cite{Geman1992}, 
\begin{align} 
\phi(x;a) = \dfrac{|x|}{1+a|x|/2},
\end{align}the logarithmic, and the arctangent penalty functions \cite{Selesnick_2014_MSC,Candes2008} are examples that satisfy Assumption \ref{theorem::assumption 1}. The rational penalty $\phi$ for $a=0.4$ is shown in Fig.~\ref{fig::phi minus absolute}.

The proximity operator of $\phi$ \cite{Combettes2007}, $\mbox{prox}_{\phi}:\mathbb{R}\to\mathbb{R}$, is defined as
\begin{align}
\label{eq::proximal of phi}
\mbox{prox}_{\phi}(y; \lambda,a) := \arg\min_{x \in \mathbb{R}} \left\lbrace \dfrac{1}{2}(y-x)^2 + \lambda\phi(x;a) \right\rbrace.
\end{align} For $\phi(x;a)$ satisfying Assumption \ref{theorem::assumption 1}, with $a < 1/\lambda$, the proximity operator is a continuous non-linear threshold function with $\lambda$ as the threshold value, i.e., $\mbox{prox}_{\phi}(y; \lambda,a) = 0, \forall |y| < \lambda$. The proximity operator of the absolute value function is the soft-thresholding function. There is a constant gap between the identity function and the soft-threshold function due to which the non-zero values are underestimated \cite{Fan2001}. On the other hand, non-convex penalty functions satisfying Assumption \ref{theorem::assumption 1} are specifically designed so that the threshold function approaches identity asymptotically. These non-convex penalty functions do not underestimate  large values.

\subsection{Convexity Condition}

\par In order to benefit from convex optimization principles in solving \eqref{eq::Cost function}, we seek to ensure $F$ in \eqref{eq::Cost function} is convex by controlling the parameter $a_i$. For later, we note the following lemma.

\newtheorem{lemma}{\bf Lemma}
\begin{lemma}
\label{theorem::Lemma 1}
Let $\phi\colon\mathbb{R}\to\mathbb{R}$ satisfy Assumption \ref{theorem::assumption 1}. The function $s\colon\mathbb{R}\to\mathbb{R}$ defined as
\begin{align}
s(x;a):= \phi(x;a) - |x|,
\end{align}is twice continuously differentiable and concave with
\begin{align} 
-a \leq s''(x;a) \leq 0.
\end{align}
\end{lemma}
\begin{IEEEproof}
Since $\phi$ and the absolute value function are twice continuously differentiable on $\mathbb{R} \setminus \lbrace 0\rbrace$, we need only show $s'(0^+) = s'(0^-)$ and $s''(0^+) = s''(0^-)$. From assumption \ref{theorem::assumption 1}, we have $\phi'(0^+) = 1$, hence $s'(0^+) = \phi'(0^+) - 1 = 0$. Again by assumption \ref{theorem::assumption 1} we have $\phi'(0^-) = -\phi'(0^+) = -1$, hence $s'(0^-) = \phi'(0^-) + 1 = 0$. Further, $s''(0^+) = \phi''(0^+)$ and $s''(0^-) = \phi''(0^-) = \phi''(0^+) = s''(0^+)$. Thus the function $s$ is twice continuously differentiable. The function $s$ is concave since $s''(x) = \phi''(x) \leq 0, \forall x \neq 0$. Using Assumption \ref{theorem::assumption 1} it follows that $-a \leq s''(x;a) \leq 0$. 
\end{IEEEproof}

Figure~\ref{fig::phi minus absolute} displays the function $s(x;a)$, which is twice continuously differentiable even though the penalty function $\phi$ is not differentiable. The following theorem states the critical value of parameter $a_i$ to ensure the convexity of  $F$ in \eqref{eq::Cost function}. 
\newtheorem{corollary}{\bf Corollary}
\newtheorem{theorem}{\bf Theorem}
\begin{theorem}
\label{theorem::Theorem 1}
Let $\phi(x; a)$ be a non-convex penalty function satisfying Assumption \ref{theorem::assumption 1} and $\vec{A}$ be a transform satisfying $\vec{A}^{T}\vec{A} = r\vec{I}$, $r>0$. The function $F:\mathbb{R}^n \to \mathbb{R}$ defined in \eqref{eq::Cost function} is strictly convex if 
\begin{align}
\label{eq::convexity condition}
0 \leq a_i < \dfrac{1}{r\lambda_i}.
\end{align}
\end{theorem}
\begin{IEEEproof}
Consider the function $G: \mathbb{R}^n \to \mathbb{R}$ defined as
\begin{align}
\label{eq::G}
G(\vec{x}) := \dfrac{1}{2}\|\vec{y}-\vec{x}\|_2^2 + \sum_{i=1}^{m}\lambda_i s([\vec{Ax}]_i; a_i). 
\end{align} Since $G$ is twice continuously differentiable (using Lemma \ref{theorem::Lemma 1}), the Hessian of $G$ is given by
\begin{align}
\nabla^2 G(\vec{x}) = \vec{I} + \vec{A}^{T}\mbox{diag}\left(\lambda_1d_1, \hdots, \lambda_md_m \right)\vec{A},
\end{align}where $d_i = s''\left( [\vec{Ax}]_i;a_i\right)$. Using \eqref{eq::Parseval frame}, we write the Hessian as
\begin{align} 
\nabla^2 G(\vec{x}) &= \vec{A}^{T}\left(\dfrac{1}{r}\vec{I} + \mbox{diag}(\lambda_1d_1,\hdots,\lambda_md_m) \right) \vec{A} \\
&= \vec{A}^{T}\mbox{diag}\left(\dfrac{1}{r} + \lambda_1d_1,\hdots, \dfrac{1}{r}+\lambda_md_m \right) \vec{A}.
\end{align}The transform $\vec{A}$ has full column rank, from \eqref{eq::Parseval frame}, hence $\nabla^2 G(\vec{x})$ is positive definite if 
\begin{align}
\label{eq::condition diagonal}
\dfrac{1}{r} + \lambda_id_i > 0, \quad i = 1,\hdots,m.
\end{align} Thus, $\nabla^2 G(\vec{x})$ is positive definite if
\begin{align}
s''([\vec{Ax}]_i;a_i) > -\dfrac{1}{r\lambda_i}.
\end{align} Using Lemma \ref{theorem::Lemma 1}, we obtain the critical value of $a_i$ to ensure the convexity of $G$, i.e.,
\begin{align}
0 \leq a_i < \dfrac{1}{r\lambda_i}.
\end{align}It is straightforward that
\begin{align}
F(\vec{x}) &= G(\vec{x}) + \sum_{i=1}^{m}\lambda_i|[\vec{Ax}]_i|.
\end{align} Thus, being a sum of a strictly convex function and a convex function, $F$ is strictly convex. 
\end{IEEEproof}

Note that if $a_i > 1/(r\lambda_i)$, then the function $G(\vec{x})$ is not convex, as the Hessian of $G(\vec{x})$ is not positive definite. As a result, $1/(r\lambda_i)$ is the critical value of $a_i$ to ensure the convexity of the function $F$. The following corollary provides a convexity condition for the situation where the same regularization parameter is applied to all coefficients.

\begin{corollary}
\label{theorem::corollary 1}
For $\lambda_i = \lambda, i = 1,\hdots,m$, the function $F$ in \eqref{eq::Cost function} is strictly convex if $0 \leq a_i < 1/(r\lambda).$ \hfill $\square$
\end{corollary} 
\par We illustrate the convexity condition using a simple example with $n=2$. We set
\begin{align}
\vec{A}^{T} = \left[\begin{array}{cccc}
1 & 1 & 1 & 1\\
1 & 1 & -1 &-1
\end{array} \right], \quad \vec{A}^{T}\vec{A} = 4\vec{I}, 
\end{align}and $\lambda_1=\lambda_2=1$. Theorem \ref{theorem::Theorem 1} states	that the function $G$ defined in \eqref{eq::G} is convex for $a_i \leq 1/4$ and non-convex for $a_i > 1/4$. 
\begin{figure}
\centering
\includegraphics[scale = 0.95]{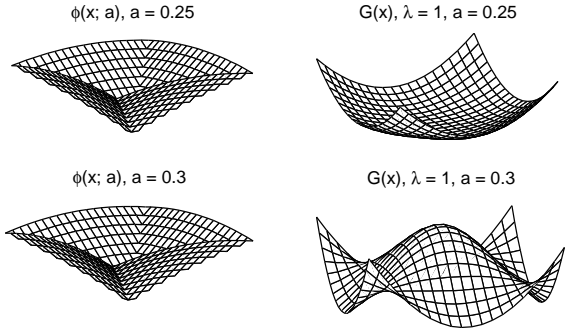}
\caption{Surface plots of the rational penalty function and the function $G$, for two different values of $a$. }
\label{fig::convexity condition}
\end{figure}
\noindent It can be seen in Fig.~\ref{fig::convexity condition} that the function $G$ is convex for $a_i = 0.25$, even though the penalty function is not convex. However, when $a_i > 0.25$, the function $G$ (hence $F$) is non-convex.

\section{Algorithm}
\label{sec::Algorithm}
\par A benefit of ensuring convexity of the objective function is that we can utilize convex optimization approaches to obtain the solution. In particular, for $\phi(x) = |x|$, the widely used methods for solving \eqref{eq::Cost function} are proximal methods \cite{Combettes2007, Combettes2010} and ADMM \cite{Boyd2010a,Goldstein2014}.

\par The convergence of ADMM to the optimum solution is guaranteed when the functions appearing in the objective function are convex \cite{Eckstein1992}. The following theorem states that ADMM can be used to solve \eqref{eq::Cost function} with guaranteed convergence, provided the augmented Lagrangian parameter $\mu$ is appropriately set. Such a condition on $\mu$ was also given in \cite{Lanza2015}. Note that $\mu$ does not affect the solution to which ADMM converges, rather the speed at which it converges. 

\begin{table}
\centering
\caption{Iterative algorithm for the solution to \eqref{eq::Cost function}.}
\begin{tabular}{@{}l@{}}
\toprule 
Input: $\vec{y}$, $\lambda_i$, $r$, $a_i$, $\mu$  \\
Initialization: $\vec{u} = 0$, $\vec{d} = 0$ \\ Repeat: \\
$\qquad$ $\vec{x} \gets \dfrac{1}{1+\mu r}\left(\vec{y} + \mu\vec{A}^{T}(\vec{u}-\vec{d}) \right)$  \\
$\qquad$ $\vec{u}_i \gets \mbox{prox}_{\phi}([\vec{Ax} + \vec{d}]_i; \lambda_i/\mu_i, a_i)$ \\
$\qquad$ $\vec{d} \gets \vec{d} - (\vec{u} - \vec{Ax})$ \\
Until convergence\\
\bottomrule
\end{tabular}
\label{table::Algorithm}
\end{table}

\begin{theorem}
Let $\phi$ satisfy Assumption \ref{theorem::assumption 1} and the transform $\vec{A}$ satisfy the Parseval frame condition \eqref{eq::Parseval frame}. Let $a_i < 1/(r_i\lambda_i)$. The iterative 
algorithm \ref{table::Algorithm} converges to the global minimum of the function $F$ in \eqref{eq::Cost function} if 
\begin{align}
\label{eq::mu condition}
\mu > 1/r.
\end{align}
\end{theorem}

\begin{IEEEproof}
We re-write the problem \eqref{eq::Cost function} using variable splitting \cite{Afonso2010} as
\begin{subequations}
\label{eq::variable split}
\begin{align}
\arg\min_\vec{u,x} &\left\lbrace \dfrac{1}{2}\|\vec{y}-\vec{x}\|_2^2 + \sum_{i=1}^{m}\lambda_i\phi\left(u_i;a_i \right) \right\rbrace \\
\mbox{s.t.}  &\quad \vec{u} = \vec{Ax}.
\end{align}
\end{subequations} The minimization is separable in $\vec{x}$ and $\vec{u}$. 
Applying ADMM to \eqref{eq::variable split} yields the following iterative procedure with the augmented Lagrangian parameter $\mu$.
\begin{subequations}
\label{eq::iterative procedure}
\begin{align}
\label{eq::subproblem in x}
\vec{x} &\gets \arg\min_{\vec{x}}\Biggl\lbrace \dfrac{1}{2}\|\vec{y} - \vec{x}\|_2^2 + \dfrac{\mu}{2}\|\vec{u}-\vec{Ax}-\vec{d}\|_2^2\Biggr\rbrace \\
\label{eq::subproblem in u}
\vec{u} &\gets \arg\min_{\vec{u}}\Biggl\lbrace \underbrace{\sum_{i=1}^{m}\lambda_i\phi\left(u_i; a_i \right) + \dfrac{\mu}{2}\|\vec{u}-\vec{Ax}-\vec{d}\|_2^2}_{R(u)} \Biggr\rbrace \\
\vec{d} &\gets \vec{d} - \left(\vec{u}-\vec{Ax} \right)
\end{align}
\end{subequations}

The sub-problem \eqref{eq::subproblem in x} for $\vec{x}$ can be solved explicitly as
\begin{align}
\label{eq::solution for x}
\vec{x} &= \left(\vec{I} + \mu\vec{A}^{T}\vec{A} \right)^{-1}\left(\vec{y} + \mu\vec{A}^{T}(\vec{u}-\vec{d})\right) \\
&= \dfrac{1}{1 + \mu r}\left(\vec{y} + \mu\vec{A}^{T}(\vec{u}-\vec{d}) \right),
\end{align}using \eqref{eq::Parseval frame}.  The sub-problem \eqref{eq::subproblem in u} for $u$ can be solved using $\mbox{prox}_{\phi}$, provided the function $R$ is convex. Consider the function $Q\colon\mathbb{R}^m\to\mathbb{R}$ defined as
\begin{align}
\label{eq::Q}
Q(\vec{u}) := \sum_{i=1}^{m}\lambda_is(u_i;a_i) + \dfrac{\mu}{2}\|\vec{u-Ax-d}\|_2^2.
\end{align} From Lemma \ref{theorem::Lemma 1} and the proof of Theorem \ref{theorem::Theorem 1}, $\nabla^2 Q(\vec{u})$ is positive definite if 
\begin{align}
s''(u_i;a_i) > \dfrac{-\mu}{\lambda_i} \quad \Rightarrow \quad \mu > a_i\lambda_i.
\end{align} Since $a_i < 1/(r\lambda_i)$, it follows that $\nabla^2Q(\vec{u})$ is positive definite if $\mu > 1/r$. Hence $Q$ is strictly convex for $\mu > 1/r$. Note that  $R(u) = Q(u) + \|u\|_1$. Hence, the function $R$, being the sum of a convex and a strictly convex function, is strictly convex. As such, the minimization problem in \eqref{eq::subproblem in u} is well-defined and its solution can be efficiently computed using the proximity operator of $\phi$ \eqref{eq::proximal of phi}, i.e.,
\begin{align}
\label{eq::solution for u}
u_i \gets \mbox{prox}_{\phi}\Bigl([Ax + d]_i; \lambda_i/\mu_i, a_i \Bigr).
\end{align}

Since $\vec{A}$ has full column rank, ADMM converges to a stationary point of the objective function (despite having a non-convex function in the objective) \cite{Magnusson2014, Wang2014a}; see also \cite{Li2014, Bolte2014, Hong2015}. Moreover, the function $F$ is strictly convex (by Theorem \ref{theorem::Theorem 1}) and the sub-problems of the ADMM are strictly convex for $\mu > 1/r$. As a result, the iterative procedure \eqref{eq::iterative procedure} converges to the global minimum of  $F$.
\end{IEEEproof}
A globally convergent algorithm based on a different splitting is presented in \cite{Ilker2015}. In that approach, the objective function is split into two functions, both of which are convex regardless of the auxillary parameter value. Hence, no parameter constraint is required to ensure convergence.

\section{Examples}
\label{sec::Examples}
\subsection{1D Signal Denoising}
We consider the problem of denoising a 1D signal that is sparse with respect to the undecimated wavelet transform (UDWT) \cite{Coifman1995}, which satisfies the condition \eqref{eq::Parseval frame} with $r=1$. In particular, we use a 4-scale UDWT with three vanishing moments. The noisy signal is generated using Wavelab (http://www-stat.stanford.edu/\%7Ewavelab/) with AWGN of $\sigma = 4.0$. We set the regularization parameters $
\lambda_j = \beta\sigma 2^{-j/2}, 1 \leqslant j \leqslant 4
$. We use the same $\lambda_j$ for all the coefficients in scale $j$. The value of $\beta$ is chosen to obtain the lowest RMSE for convex and non-convex regularization respectively. To maximally induce sparsity we set $a_i = 1/\lambda_i$. For the 1D signal denoising example, we use the non-convex arctangent penalty and its corresponding threshold function \cite{Selesnick_2014_MSC}. For comparison we use reweighted $\ell_1$ minimization \cite{Candes2008}, with $\beta$ chosen in order to obtain the lowest RMSE. 

\begin{figure}[t!]
\centering
\includegraphics[]{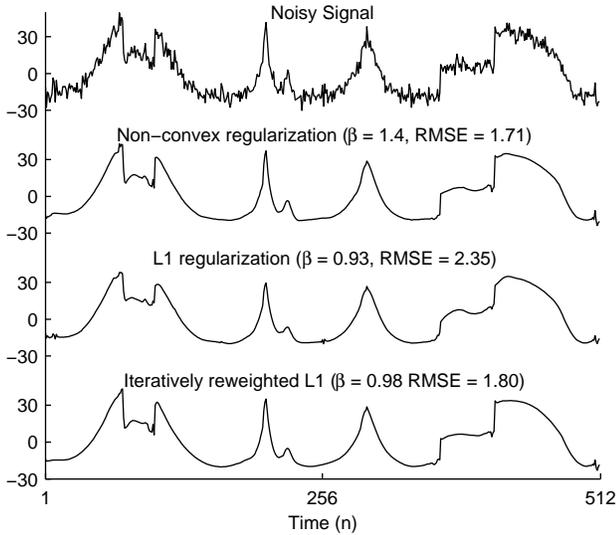}	
\caption{1D denoising example. Non-convex regularization yields lower RMSE than convex regularization.}
\label{fig::1D example}
\end{figure}

\begin{figure}[t!]
\centering
\includegraphics[]{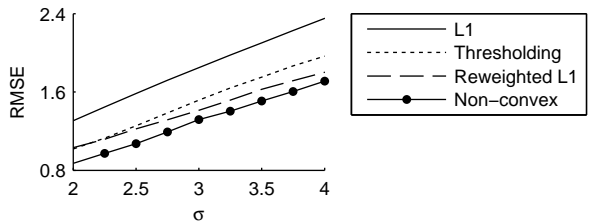}
\caption{RMSE values as a function of the noise level $\sigma$ for the 1D signal denoising example.}
\label{fig::RMSE comparison}
\end{figure}

Figure~\ref{fig::1D example} shows that the denoised signal obtained using non-convex regularization has the lowest RMSE and preserves the discontinuities. Further, the peaks are less attenuated using non-convex regularization in comparison with $\ell_1$ norm regularization. 

For further comparison, we generate the noisy signal in Fig.~\ref{fig::1D example} for $1 \leqslant \sigma \leqslant 4$, and denoise it with non-convex and convex regularization. We also denoise the noisy signal by direct non-linear thresholding of the noisy wavelet coefficients and by reweighted $\ell_1$ minimization. We use the same $\beta$ values as in Fig.~\ref{fig::1D example}. The value of $\beta$ for direct non-linear thresholding is also chosen to obtain the lowest RMSE. As seen in Fig.~\ref{fig::RMSE comparison}, the non-convex regularization outperforms the three methods by giving the lowest RMSE. The RMSE values are obtained by averaging over 15 realizations for each $\sigma$.

\subsection{2D Image Denoising}
We consider the problem of denoising a 2D image corrupted with AWGN. We use the 2D dual-tree complex wavelet transform (DT-CWT) \cite{Selesnick2005}, which is 4-times expansive and satisfies \eqref{eq::Parseval frame} with $r = 1$. The noisy `peppers' image has peak signal-to-noise ratio (PSNR) value of 14.6 dB. We use the same $\lambda$ for all the sub-bands. As in the previous example, we set the value of $\lambda$ for each case (convex and non-convex) as a constant multiple of $\sigma$ that gives the highest PSNR. 

\begin{figure}[t!]
\centering
\includegraphics[]{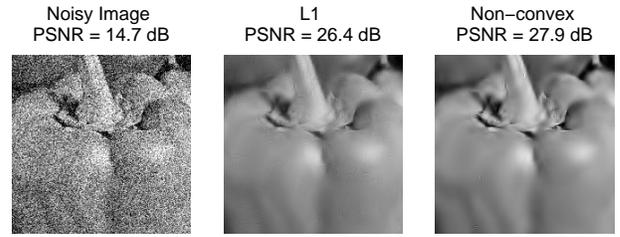}
\caption{Image denoising. Wavelet artifacts are more prominent when using $\ell_1$ norm regularization.}
\label{fig::2D example}
\end{figure}

\begin{figure}
\centering
\includegraphics[]{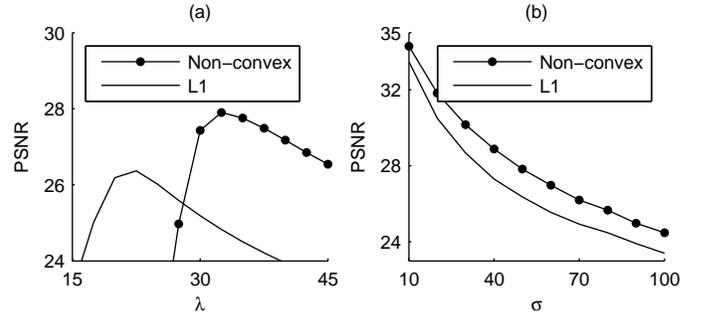}
\caption{Relative performance of convex and non-convex regularization for image denoising. (a) PSNR as a function of $\lambda$. (b) PSNR as a function of $\sigma$.}
\label{fig::PSNR values}
\end{figure}

Figure~\ref{fig::2D example} shows that the denoised image (non-convex case) contains fewer wavelet artifacts and has a higher PSNR. Figure.~\ref{fig::PSNR values}(a) shows the PSNR values (convex and non-convex) for different values of $\lambda$. To further assess the performance of tight-frame non-convex regularization, we realize several noisy `peppers' images with $10\leqslant \sigma \leqslant 100$. As in the case of the 1D signal denoising, Fig.~\ref{fig::PSNR values} shows that non-convex regularization offers higher PSNR across different noise-levels.

\section{Conclusion}

This letter considers the problem of signal denoising using a sparse tight-frame analysis prior. We propose the use of parameterized non-convex regularizers to maximally induce sparsity while maintaining the convexity of the total problem. The convexity of the objective function is ensured by restricting the parameter $a$ of the non-convex regularizer. We use ADMM to obtain the solution to the convex objective function (consisting of a non-convex regularizer), and guarantee its convergence to the global optimum, provided the augmented Lagrangian parameter $\mu$, satisfies $\mu > 1/r$. The proposed method outperforms the $\ell_1$ norm regularization and reweighted $\ell_1$ minimization methods for signal denoising.

\end{document}